\documentclass[runningheads]{llncs}

\usepackage{graphicx}
\usepackage{graphics}
\usepackage{listings}
\usepackage{todonotes} 
\usepackage{color}
\usepackage{courier}
\usepackage{booktabs}
\usepackage{float}
\usepackage{url}
\usepackage[utf8]{inputenc}

\definecolor{lightBlue}{RGB}{0, 153, 255}
\definecolor{lightRed}{RGB}{204, 0, 255}

\makeatletter
\newcommand*{\centerfloat}{
  \parindent \z@
  \leftskip \z@ \@plus 1fil \@minus \textwidth
  \rightskip\leftskip
  \parfillskip \z@skip}
\makeatother

\makeatletter
\DeclareRobustCommand\ttfamily
        {\not@math@alphabet\ttfamily\mathtt
         \fontfamily\ttdefault\selectfont\hyphenchar\font=-1\relax}
\makeatother
\DeclareTextFontCommand{\mytexttt}{\ttfamily\hyphenchar\font=45\relax}

\lstdefinelanguage{Manchester}
{
    sensitive = true,
    keywords = [1]{Class, EquivalentTo, SubClassOf},
    morekeywords = [2]{and, or},
    morekeywords = [3]{some, value},
    keywordstyle=[2]\textbf,
    keywordstyle=[2]\color{lightBlue}\textbf,
    keywordstyle=[3]\color{lightRed}\textbf,
    morestring=[b]"
}

\begin{document}

\title{A Semantic Framework for Enabling Radio Spectrum Policy Management and Evaluation\thanks{Approved for public release (reference number: 88ABW-2020-1535).}}
\titlerunning{DSA Policy Framework}

\author{Henrique Santos\inst{1}\orcidID{0000-0002-2110-6416}
\and Alice Mulvehill\inst{1,3}\orcidID{0000-0003-2981-0675}
\and John S. Erickson\inst{1,2}\orcidID{0000-0003-3078-4566}
\and James P. McCusker\inst{1}\orcidID{0000-0003-1085-6059}
\and Minor Gordon\inst{1}\orcidID{0000-0003-1928-4130}
\and Owen Xie\inst{1}\orcidID{0000-0002-8689-3385}
\and Samuel Stouffer\inst{1}\orcidID{0000-0002-8929-5989}
\and Gerard Capraro\inst{4}\orcidID{0000-0002-1836-9231}
\and Alex Pidwerbetsky\inst{5}\orcidID{0000-0001-7436-8105}
\and John Burgess\inst{5}\orcidID{0000-0001-9539-6922}
\and Allan Berlinsky\inst{5}\orcidID{0000-0001-5314-7178}
\and Kurt Turck\inst{6}\orcidID{0000-0001-9224-0164}
\and Jonathan Ashdown\inst{6}\orcidID{0000-0001-7202-1095}
\and Deborah L. McGuinness\inst{1}\orcidID{0000-0001-7037-4567}
}

\authorrunning{H. Santos et al.}

\institute{Tetherless World Constellation, Rensselaer Polytechnic Institute, Troy NY, USA \and
The Rensselaer Institute for Data Exploration and Applications, Troy NY, USA \and
Memory Based Research LLC, Pittsburgh PA, USA \and
Capraro Technologies Inc., Utica NY, USA \and
LGS Labs, CACI International Inc., Florham Park NJ, USA \and
Air Force Research Laboratory, Rome NY, USA\\
\email{\{oliveh,mulvea,erickj4,mccusj2,gordom6,xieo,stoufs\}@rpi.edu, dlm@cs.rpi.edu, gcapraro@caprarotechnologies.com, \{a.pidwerbetsky,john.burgess,allan.berlinsky\}@caci.com, \{kurt.turck,jonathan.ashdown\}@us.af.mil}
}

\maketitle

\begin{abstract}
Because radio spectrum is a finite resource, its usage and sharing is regulated by government agencies. These agencies define policies to manage spectrum allocation and assignment across multiple organizations, systems, and devices. With more portions of the radio spectrum being licensed for commercial use, the importance of providing an increased level of automation when evaluating such policies becomes crucial for the efficiency and efficacy of spectrum management. We introduce our Dynamic Spectrum Access Policy Framework for supporting the United States government's mission to enable both federal and non-federal entities to compatibly utilize available spectrum. The DSA Policy Framework acts as a machine-readable policy repository providing policy management features and spectrum access request evaluation. The framework utilizes a novel policy representation using OWL and PROV-O along with a domain-specific reasoning implementation that mixes GeoSPARQL, OWL reasoning, and knowledge graph traversal to evaluate incoming spectrum access requests and explain how applicable policies were used. The framework is currently being used to support live, over-the-air field exercises involving a diverse set of federal and commercial radios, as a component of a prototype spectrum management system.

\end{abstract}

\section{Introduction}

Usable radio spectrum is becoming crowded\footnote{\url{http://bit.ly/FCC_AWS}} as an increasing number of services, both commercial and governmental, rely on wireless communications to operate. Techniques known as Dynamic Spectrum Access (DSA)~\cite{zhao_survey_2007} have been extensively researched as a way of promoting more efficient methods for sharing the radio spectrum among distinct organizations and their respective devices, while adhering to regulations.

In the United States, spectrum is managed by agencies that include the National Telecommunications and Information Administration\footnote{\url{http://ntia.doc.gov}} (NTIA) and the Federal Communications Commission\footnote{\url{http://fcc.gov}} (FCC). The NTIA publishes revised versions of its Manual of Regulations and Procedures for Federal Radio Frequency Management\footnote{\url{http://bit.ly/NTIA_Redbook}} (commonly referred to as the NTIA Redbook) which is a compilation of regulatory policies that define the conditions that non-US, as well as federal and non-federal US, organizations, systems, and devices must satisfy in order to compatibly share radio spectrum while minimizing interference.

With the advent of 5G,\footnote{\url{http://bit.ly/FCC_5G}} more parts of the radio spectrum are being licensed for commercial usage and, with the increased availability of cognitive radios~\cite{zhang_dynamic_2008} (devices that are able to automatically adjust their operating frequency), the importance of providing an increased level of automation when evaluating spectrum policies becomes crucial for the sustainability of spectrum management. This issue is currently under investigation by the National Spectrum Consortium\footnote{\url{http://nationalspectrumconsortium.org}} (NSC), a research and development organization that incubates new technologies to tackle challenges about radio spectrum management and utilization.

In this paper, we describe the Dynamic Spectrum Access Policy Framework (DSA Policy Framework). The DSA Policy Framework supports the management of machine-readable radio spectrum usage policies and provides a request evaluation interface that is able to reason about the policies and generate permit or deny results to spectrum access requests. This is accomplished via the utilization of a novel policy representation based on semantic web standards OWL and PROV-O that encodes its rules in an ontology. This ontology, combined with background knowledge originating from a number of relevant select sources, is stored in a Knowledge Graph that is used by a domain-specific reasoning implementation that mixes GeoSPARQL~\cite{perry_ogc_2012}, OWL reasoning, and knowledge graph traversal to evaluate policies that are applicable to spectrum access requests. Effects (Permit/Deny/Permit with Obligations) are assigned and explanations are provided to justify why a particular request was permitted or denied access to the requested frequency or frequency range.

\section{Sharing the radio spectrum}

Radio spectrum policies specify how available spectrum should be used and shared.  The applicability of existing policies must be checked for a variety of activities that use spectrum in many settings including various different types of requesters (e.g. systems and devices). For example, training exercises will request spectrum usage for a certain time frame and geographic region for potentially hundreds of radios with a wide variety of capabilities. During a training exercise, local policies are typically created to manage the  spectrum that radios used in the exercise will require, and to minimize interference between federal and commercial (non-federal) radios that may be operating in the same area and within the same frequency range. Throughout this paper the following definitions are used:

\begin{itemize}
    \item \emph{High-level policies}: Policies as documented by authoritative agencies, including NTIA and FCC. These policies are not prone to change in the short term, although they may evolve when new versions of documents are released.
    \item \emph{Local policies}: Specializations of high-level policies. These are created to locally manage the spectrum requests of various devices that want to operate in specific locations and/or for specific time periods.
    \item \emph{Sub-policies}: A sub-section of a policy, sometimes referred to as ``provisions'' in NTIA Redbook policies. 
    \item \emph{Spectrum manager}: The role of a human who is responsible for managing policies. The spectrum manager verifies if existing policies are sufficient to support some activity and creates local policies to accommodate specific spectrum requirements.
    \item \emph{Spectrum system}: This role represents some external system that is being used to generate spectrum requests on behalf of entities that require the use of a specific frequency or set of frequencies.
\end{itemize}

\section{Dynamic Spectrum Access Policy Framework}

The DSA Policy Framework supports the following objectives:

\begin{itemize}
    \item Serve as a centralized machine-readable radio spectrum policy repository.
    \item Provide policy management features (including creation and customization) for a wide range of radio spectrum domain users.
    \item Use machine-readable policies as a basis for automatically evaluating radio spectrum access requests. 
\end{itemize}

\begin{figure}
    \centering
    \includegraphics[width=\textwidth]{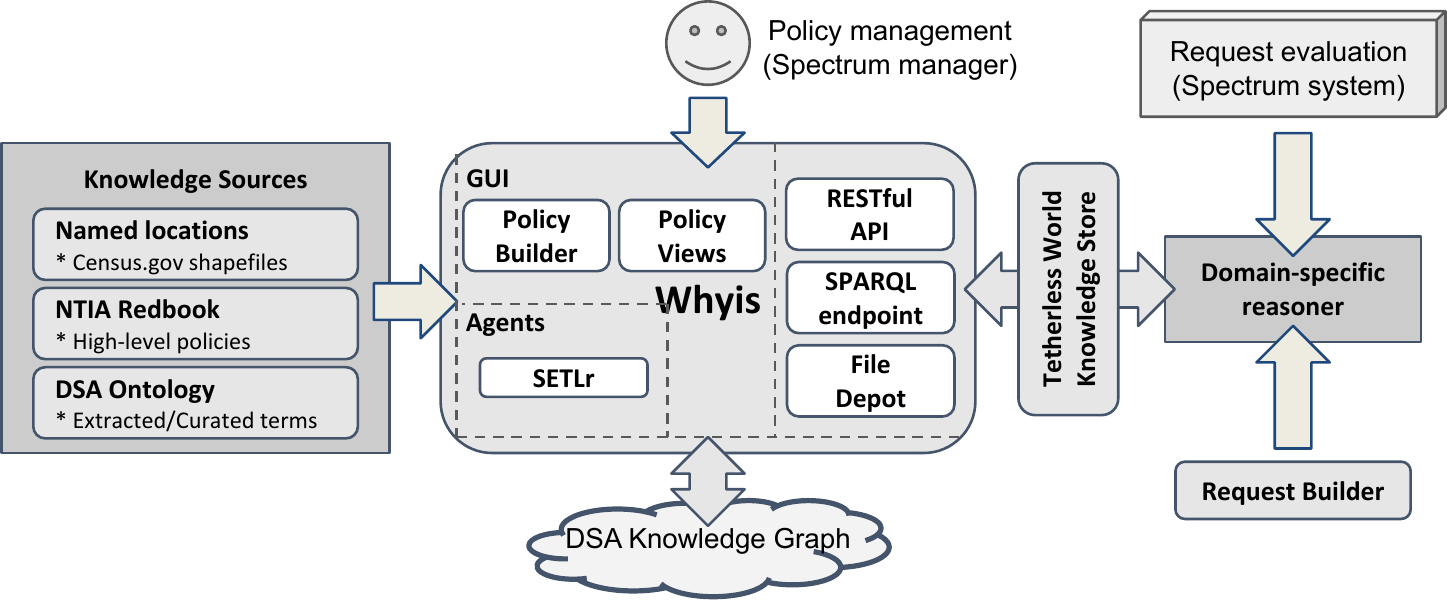}
    \caption{DSA Policy Framework architecture}
    \label{fig:architecture}
\end{figure}

As shown on the top of Figure~\ref{fig:architecture}, the DSA Policy Framework provides two major functions: \emph{Policy Management} and \emph{Request Evaluation}. 
\emph{Policy Management} enables the spectrum manager to create local policies by referencing relevant higher level policies and adding customization. \emph{Request Evaluation} utilizes all policies to automatically process spectrum requests. The results include references to any policy that was involved in the evaluation. As policies evolve, the underlying knowledge representation evolves, enabling the \emph{request evaluation} module to use the most current policy information to reason and assign effects (permit, deny, permit with obligation) to spectrum requests. Spectrum managers can verify evaluation results in the presence of newly created policies through the \emph{Request Builder} tool.

To support the DSA Policy Framework infrastructure we leveraged Whyis~\cite{mccusker_whyis_2018}, a nanopublication-based knowledge graph publishing, management, and analysis framework. Whyis enables the creation of domain and data-driven knowledge graphs. The DSA Policy Framework takes advantage of the use of nanopublications~\cite{groth_anatomy_2010} in Whyis, which allows it to incrementally evolve knowledge graphs while providing provenance-based justifications and publication credit for each piece of knowledge in the graph. This is particularly useful for the spectrum domain because policies do change and new policies can be created, potentially triggered by multiple sources. The DSA Policy Framework also makes use of the SETLr~\cite{mccusker_setlr_2018} Whyis agent, which enables the conversion of a number of the identified knowledge sources or derivatives to the Resource Description Framework (RDF), thereby bootstrapping the DSA Knowledge Graph.

\subsection{Knowledge Sources}

In order to support the creation and interpretation of machine-readable radio spectrum policies, information from several sources was mined and incorporated into a Knowledge Graph. Spectrum policies and term definitions were obtained from the NTIA Redbook, an IEEE Standard Definitions and Concepts document~\cite{noauthor_ieee_2008}, and from FCC 14-31~\cite{noauthor_federal_2014} (a FCC policy publication). Specific schemas associated with the spectrum domain were obtained from the Standard Spectrum Resource Format (SSRF)~\cite{noauthor_standard_2014}. Finally, information about geographical locations was obtained from the Census.gov shapefiles.\footnote{\url{http://bit.ly/Census_shapefiles}}

During the \emph{policy capture} process, Rensselaer Polytechnic Institute (RPI) collaborated with DSA domain experts from Capraro Technologies Inc. and LGS Labs of CACI International Inc. to select and analyze English text-based policies from the NTIA Redbook and from various FCC documents. In order to be used by the DSA Policy Framework, the English text was converted into a different representation, and many of the terms used in the English text were incorporated into a domain ontology. During this process, we observed that the text for many policies is logically equivalent to a conditional expression e.g., \emph{IF} some device wants to use a frequency in a particular frequency range \emph{AND} at a particular location, \emph{THEN} it is either \emph{PERMITTED} or \emph{DENIED}.

\begin{figure}
    \centering
    \includegraphics[width=\textwidth]{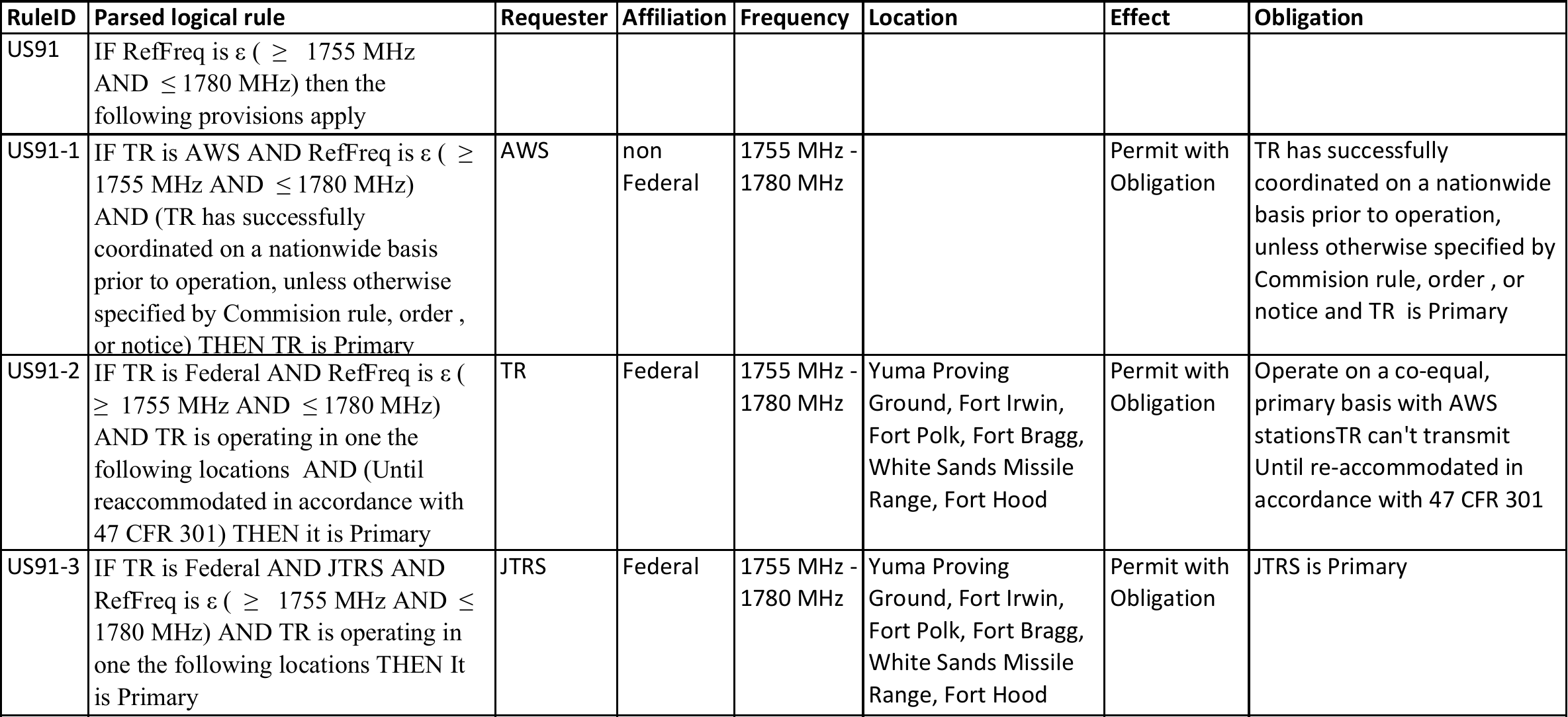}
    \caption{Spreadsheet excerpt showing the NTIA Redbook US91 policy capture}
    \label{fig:policy-capture}
\end{figure}

More complex policies contain a set of conditional expressions, with each conditional expression focused on a particular request attribute, e.g., a request device type, frequency, frequency range.  The spreadsheet displayed in Figure~\ref{fig:policy-capture} contains an example of a complex policy from the NTIA Redbook called US91. Due to space constraints, we have omitted some of the sub-policies for US91 and the  columns that document policy metadata and provenance, including the original  text, source document, URL, and page number.

US91 regulates the usage of the 1755-1780MHz frequency range and is an example of a spectrum range that must be efficiently shared by both federal and non-federal devices. Because spectrum usage can vary by device, affiliation and location, we decompose the policy into several sub-policies (US91-1, US91-2, and US91-3). A \emph{parsed logical rule} is manually created for each policy by a domain expert. The elements of the logical rule are further expressed as attribute-value pairs, e.g., \mytexttt{Requester = AWS}. The attribute (column) names map into the following elements of the \emph{policy logical expression} that is used by the framework:

\begin{itemize}
    \item \emph{Requester:} the device requesting access
    \item \emph{Affiliation:} the affiliation of the requester (Federal, Non-Federal)
    \item \emph{Frequency:} the frequency range or single frequency being requested
    \item \emph{Location:} location(s) where the policy is applicable
    \item \emph{Effect:} the effect a policy yields, if the rule is satisfied (Permit, Deny, Permit with Obligations)
    \item \emph{Obligations:} the list of obligations the requester needs to comply with in order to be permitted
\end{itemize}

The framework utilizes several ontologies to support policy administration and spectrum request processing, including a domain ontology called the DSA Ontology. DSA ontology terms were collected during policy capture and/or derived from the NTIA Redbook, IEEE Standards, SSRF, or other domain source. All terms were curated by an ontology developer and linked to external ontologies including PROV-O and the Semanticscience Integrated Ontology (SIO)~\cite{dumontier_semanticscience_2014}.

\subsection{Representing radio spectrum requests}

\begin{figure}
    \centering
    \includegraphics[width=\textwidth]{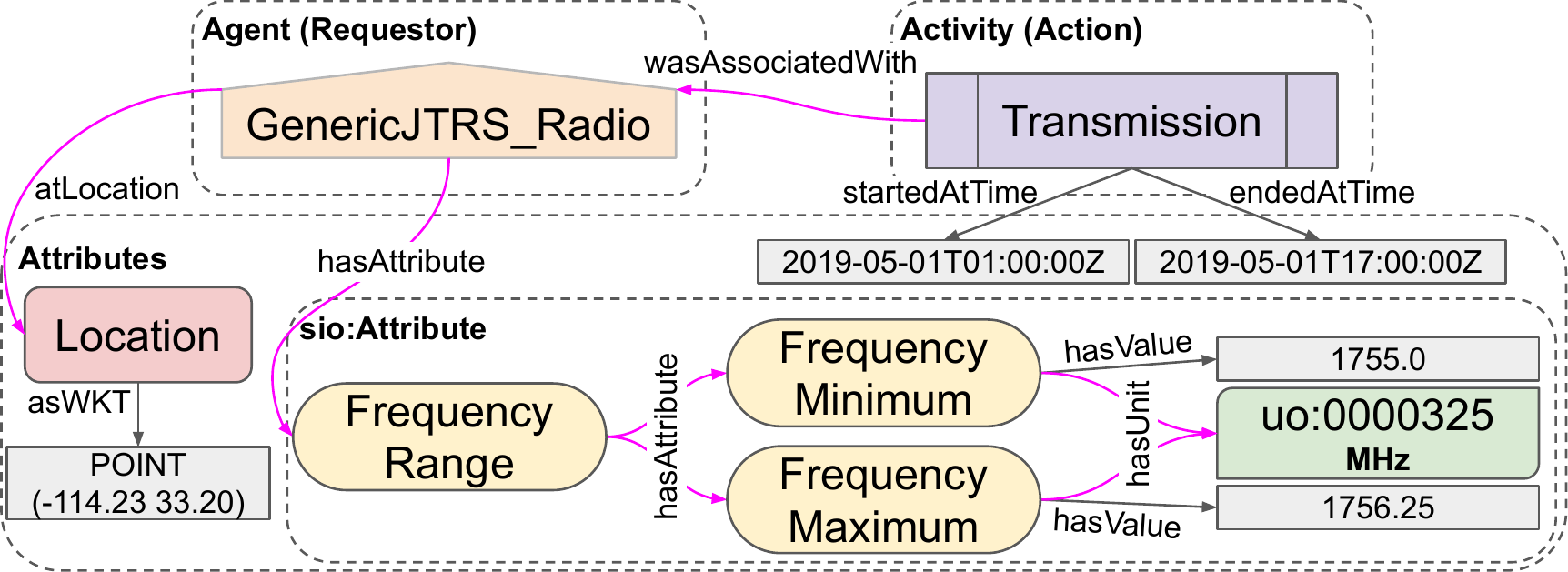}
    \caption{The DSA request model and sample request}
    \label{fig:request-prov}
\end{figure}

Figure~\ref{fig:request-prov} shows the DSA request model and a sample request. The model is based on the World Wide Web Consortium's recommended standard for provenance on the web (PROV). The modeling of requests as activities and agents was influenced by the policy attributes described as columns in the policy capture spreadsheet (shown in Figure~\ref{fig:policy-capture}) and extended to include the action associated with the requester in a spectrum request (currently we represent only the \mytexttt{Transmission} action). In the model, the requester (\mytexttt{prov:Agent}) is linked with an action \mytexttt{prov:Activity} using the \mytexttt{prov:wasAssociatedWith} predicate. The location attribute is represented as \mytexttt{prov:Location} and linked to the requester using the \mytexttt{prov:atLocation} predicate. The time attribute, which describes when the request action is to start and end, is represented using the literal data type \mytexttt{xsd:dateTime} and linked to the action using the \mytexttt{prov:startedAtTime} and \mytexttt{prov:endedAtTime} predicates.

For attributes not natively supported by PROV, we use SIO, which enables us to model objects and their attributes (roles, measurement values) with the use of the \mytexttt{sio:Attribute} class and \mytexttt{sio:hasAttribute}~/\mytexttt{sio:hasValue}~/\mytexttt{sio:hasUnit} predicates. In the DSA request model, the attributes Frequency, Frequency Range, and Affiliation are represented as \mytexttt{sio:Attribute} and linked to the requester using \mytexttt{sio:hasAttribute}. For attributes that can assume a value (currently only Frequencies), literal values (\mytexttt{sio:hasValue}) and units of measurement (\mytexttt{sio:hasUnit}) are used to express the quantification of an attribute as a specified unit from the Units of Measurement Ontology (UO)~\cite{gkoutos_units_2012}. In Figure~\ref{fig:request-prov}, a sample radio spectrum request instance is shown, where the requester is a Generic Joint Tactical Radio System (JTRS) radio. The device is physically located at a location that is defined by the Well-Known Text (WKT)~\cite{noauthor_isoiec_nodate} string \mytexttt{POINT(-114.23 33.20)} and is requesting access to the frequency range \mytexttt{1755 - 1756.25 MHz} using the \mytexttt{FrequencyRange} attribute, which is described by the composition of the \mytexttt{FrequencyMinimum} and \mytexttt{FrequencyMaximum} attributes, with their respective value and unit.

\subsection{Representing radio spectrum policies}

The DSA policy model was created to enable (1) the unambiguous representation of rules as parsed during \emph{policy capture}, (2) the reuse of an existing policy’s rules for creating local policies, and (3) the implementation of request evaluation capabilities. The model is based on OWL, encoding policy rules as restrictions in OWL classes that represent policies. The OWL restrictions are constructed on the RDF properties of the DSA request model, presented in the previous subsection, while constraining their ranges to the expected literal value or class, as dictated by the policy rule being expressed. To demonstrate this, Listing~\ref{lst-policy-manchester} shows the OWL expression in Manchester syntax of an excerpt of the NTIA US91 policy that describes how a requester that is an example of a radio, categorized as JTRS, can use the spectrum regulated by US91 (1755-1780 MHz) in specific locations (US91-3 sub-policy in Figure~\ref{fig:policy-capture}).

The model advocates for the creation of an OWL class for each rule that composes the complete policy rule expression. In the example, the OWL class \mytexttt{US91} has the restriction on the frequency range attribute, with the minimum value \mytexttt{1755} and maximum value \mytexttt{1780} (lines 4-11). This class is a subclass of \mytexttt{Transmission} (line 13), which is the action this policy regulates. In lines 15-20, a class \mytexttt{US91-3} restricts the requester to a \mytexttt{JointTacticalRadioSystem} and appends that rule to the rules of its super-class \mytexttt{US91}. The last class of this policy expression is \mytexttt{US91-3.1}, which encodes the location rule as a restriction on the \mytexttt{atLocation} property (lines 24-25), constraining it to a location class. When the composition of rules expressed by this last class is evaluated to be true, the policy should assign the permit effect. This is expressed by asserting  \mytexttt{US91-3.1} as a subclass of \mytexttt{Permit} (line 27).

\lstset{language=Manchester, basicstyle=\ttfamily\fontsize{9}{10}\selectfont, columns=fullflexible, xleftmargin=5mm, framexleftmargin=5mm, numbers=left, stepnumber=1, breaklines=true, breakatwhitespace=false, numberstyle=\ttfamily\fontsize{9}{10}\selectfont, numbersep=5pt, tabsize=2, frame=lines, captionpos=b, caption={OWL expression of part of the US91 policy in Manchester syntax}, label=lst-policy-manchester}
\begin{lstlisting}
Class: US91
  EquivalentTo:
    Transmission and
    (wasAssociatedWith some (hasAttribute some
      (FrequencyRange
       and (hasAttribute some
             (FrequencyMaximum and
               (hasValue some xsd:float[<= 1780.0f])))
       and (hasAttribute some
             (FrequencyMinimum and
               (hasValue some xsd:float[>= 1755.0f]))))))
  SubClassOf:
    Transmission
    
Class: US91-3
  EquivalentTo:
    US91 and
    (wasAssociatedWith some JointTacticalRadioSystem)
  SubClassOf:
    US91

Class: US91-3.1
  EquivalentTo:
    US91-3 and (wasAssociatedWith some
                    (atLocation some US91-3.1_Location))
  SubClassOf: 
    Permit, US91-3
\end{lstlisting}

Many of the NTIA policies, including US91, contain location rules and, often, these rules are written in terms of location lists. To represent this, we defined OWL classes for the lists to be used in conjunction with the OWL expression of policies. Listing~\ref{lst-location-manchester} shows the definition of the \mytexttt{US91-3.1\_Location} class, which is used in the \mytexttt{US91-3.1} restriction, as previously shown. We have used the GeoSPARQL predicate \mytexttt{sfWithin} in conjunction with OWL unions to express that the rule is satisfied if the location is specified in the list (lines 3-8). In this example, we leverage location information we imported from Census.gov shapes for US Federal locations.

\lstset{language=Manchester, basicstyle=\ttfamily\fontsize{9}{10}\selectfont, columns=fullflexible, xleftmargin=5mm, framexleftmargin=5mm, numbers=left, stepnumber=1, breaklines=true, breakatwhitespace=false, numberstyle=\ttfamily\fontsize{9}{10}\selectfont, numbersep=5pt, tabsize=2, frame=lines, captionpos=b, caption={OWL expression of a location list in Manchester syntax}, label=lst-location-manchester}
\begin{lstlisting}
Class: US91-3.1_Location
  EquivalentTo:
    (sfWithin value White_Sands_Missile_Range) or
    (sfWithin value Ft_Irwin) or
    (sfWithin value Yuma_Proving_Ground) or
    (sfWithin value Ft_Polk) or
    (sfWithin value Ft_Bragg) or
    (sfWithin value Ft_Hood)
  SubClassOf:
    Location
\end{lstlisting}

The creation of the OWL class hierarchy (and, therefore, the incremental addition of rules) maximizes the reuse of rules when spectrum managers create local policies. To illustrate this claim, Listing~\ref{lst-local-manchester} shows a sample local policy which modifies the existing US91-3 sub-policy to \mytexttt{Deny} requests in a specific time window. Lines 4-5 contain the time restrictions on the PROV-O predicates. Local policies can have an explicit precedence level, as shown in line 7.

\lstset{language=Manchester, basicstyle=\ttfamily\fontsize{9}{10}\selectfont, columns=fullflexible, xleftmargin=5mm, framexleftmargin=5mm, numbers=left, stepnumber=1, breaklines=true, breakatwhitespace=false, numberstyle=\ttfamily\fontsize{9}{10}\selectfont, numbersep=5pt, tabsize=2, frame=lines, captionpos=b, caption={OWL expression of a local policy in Manchester syntax}, label=lst-local-manchester}
\begin{lstlisting}
Class: US91-3.1-Local
  EquivalentTo:
    US91-3.1 and
    endedAtTime some xsd:dateTime[>=2019-10-01T11:00:00Z] and
    startedAtTime some xsd:dateTime[<=2019-10-01T17:00:00Z]
  SubClassOf: 
    Deny, Priority_1, US91-3.1
\end{lstlisting}

\subsection{Policy management}

The DSA Policy Framework provides a web interface to allow spectrum managers to have a comprehensive understanding of the DSA Knowledge Graph, including policies, locations, and entities in the DSA Ontology. It leverages Whyis default ``views'' with some extensions for supporting a domain-specific display of pieces of the Knowledge Graph. The structure and content of the interface are driven by the DSA Knowledge Graph, which ensures that it displays relevant and contextualized information and features.

The \emph{Policy Faceted Browser} that allows a user to quickly find policies based on the selection of attribute values. Spectrum managers can, for instance, find policies applicable to a list of select locations or policies applicable to a specific device, or even to a combination of both. This is accomplished by domain-specific SPARQL queries that retrieve and group attributes from the policy's OWL structure. The user can view details of a policy or reuse a policy's rules.

The \emph{Policy Detail} view provides a display of policy metadata, including name, original text and identifier, and a human-readable version of the policy encoded rules. If the policy specifies locations, those locations will be displayed on a map.

The \emph{Policy Builder} view can be used to build a policy from scratch or to create local policies by reusing existing policies' rules. The Policy Builder leverages the DSA Knowledge Graph to provide user support during policy creation. For instance, if a user wants to create a rule for an specific device, the Builder will display known devices as represented in the KG. More than that, the Builder ``understands'' the semantics of the rule which means that if the user wants to add a frequency range rule, for instance, the Builder will prompt the user to enter both lower and upper bound values. Users can also set policy effects, precedence, and obligations. In the back end, the policy is converted to the DSA policy model in OWL and stored as a new piece of knowledge in the DSA Knowledge Graph.

Knowledge curation is a time-consuming task and it might impact the pace at which updated knowledge becomes available for use when creating new policies. To overcome this, the DSA Policy Framework supports policy additions that refer to terms not currently in the ontology, by allowing input of new terms along with the annotation that those terms need in order to be curated by the appropriate ontology owner. This allows a spectrum manager to input and test a new policy without having to wait for an ontology update first.

\subsection{Request evaluation: Domain-specific reasoning}
\label{sec:policy-evaluation}

To enable the evaluation of spectrum requests against policies, we have implemented a domain-specific reasoner that combines various Semantic Web computational approaches to assign \mytexttt{Permit/Deny} effects to requests, while fulfilling requirements, including geographical reasoning, policy precedence evaluation, and explanations for denied requests. The domain-specific reasoner follows a four-phase pipeline, with a set of requests as input and the assigned effect, a list of obligations, and a list of reasons for each request as output. The reasoner initiates by creating an in-memory RDF graph originated by the merge of the request RDF graph and the DSA Knowledge Graph.

Next, in the \textbf{geographical reasoning phase}, the reasoner elicits the geographical relationships among the requests' WKT locations, and the named locations present in the DSA KG, by using GeoSPARQL to infer triples like \mytexttt{:req\_location geo:sfWithin :NAMED\_LOCATION}. The inferred triples are then asserted back into the graph.

In the \textbf{OWL reasoning phase}, the reasoner makes use of the  HermiT OWL reasoner~\cite{glimm_hermit_2014} to perform Description Logic (DL) reasoning over the updated graph. The domain-specific implementation relies on the DL services of:

\begin{itemize}
    \item \emph{Classification} for computing all subclass relationships, allowing the inferred class hierarchy to be leveraged when querying policy Effect and Precedence.
    \item \emph{Realization} of computing classes (policies) that individuals (requests) belong to. ``Belonging'' to a class means that an individual satisfies the constraints of that class; realization can be understood as determining \emph{policy applicability}.
    \item \emph{DL Query} for retrieving the individuals that were determined to belong to specific classes.
\end{itemize}

Using the list of applicable policies retrieved from HermiT, the domain-specific reasoner decides precedence, in the \textbf{precedence evaluation phase}, by a simple evaluation of which policy has the highest precedence level. Policies with no explicit precedence are assumed to have the lowest precedence. The highest precedence policy effect is then assigned to the request. The reasoner follows these heuristics to explain the assignment of a \mytexttt{Deny} effect to a request, in the \textbf{explanation phase}:

\begin{itemize}
    \item Requests can be denied by a specific applicable policy with a \mytexttt{Deny} effect. In this case, the ``reason'' for the denial is the identity of this policy and the specification of what attributes of the request fulfill the rules contained within. The reasoner retrieves the policy's rules and presents them as reasons for denial.
    \item Requests can be denied due to a lack of a policy with a \mytexttt{Permit} effect (i.e. there is no applicable policy with either a \mytexttt{Deny} or \mytexttt{Permit} effect). In these cases the reasoner finds the rules that the request did not satisfy in order to be assigned a \mytexttt{Permit}. The reasoner calculates the paths in the OWL class hierarchy from those policies that the request was reasoned to belong to, to the policies that would result in a \mytexttt{Permit}. These paths contain classes that the request was determined not to belong to. Unfulfilled rules for each policy found to be in the path are retrieved and presented as reasons for the \mytexttt{Deny}.
    
\end{itemize}

As as example, the request in Figure~\ref{fig:request-prov} would ultimately be determined to belong to the \mytexttt{US91}, \mytexttt{US91-3}, and \mytexttt{US91-3.1} classes displayed in Listing~\ref{lst-policy-manchester} and, therefore, assigned the \mytexttt{Permit} effect. Nevertheless, when the local policy shown in Listing~\ref{lst-local-manchester} exists in the graph, the request is then reasoned to belong to the \mytexttt{US91-3.1-Local} class as well. Since this local policy contains a time constraint rule with a higher precedence, the reasoner assigns the effect to be a \mytexttt{Deny} and returns the reason, ``the request is in a prohibited time window.''

Conversely, if the request is modified to a different location outside the locations expressed in Listing~\ref{lst-location-manchester}, it would be reasoned to belong only to \mytexttt{US91} and \mytexttt{US91-3} classes, based on the frequency range and requester attributes of the request. These applicable classes don't express an \emph{explicit} \mytexttt{Permit} or \mytexttt{Deny} effect, so the reasoner must default to \mytexttt{Deny}. The reasoner then calculates the path to the class \mytexttt{US91-3.1} that would \mytexttt{Permit} it \emph{if} the conditions had been met, and retrieves the rules of each class in the path (in this case, only the rules of the \mytexttt{US91-3.1} class) and returns the reason, ``the request is not in a permitted location.''

\section{Evaluation of the Framework}

In order to evaluate the DSA Policy Framework’s semantic representation of the domain, we identified several fundamental spectrum policy constructs. They are displayed in bold in the first column of Table~\ref{tab:evaluation}. For each of them, we worked with domain experts to identify the elements that were required in order to effectively represent policies and support request evaluation. The table contains columns for Policy Representation (high-level and local policies) and Request Evaluation. ``Yes'' in the columns indicates that the policy construct is either \emph{Relevant} or it has been fully addressed and \emph{Implemented}. ``Partial'' indicates that the current implementation meets a simplified version of the requirement, while  ``uc'' means that the construct element is currently under consideration. 

\begin{table}[ht!]
  \centerfloat
  \resizebox{.8\textwidth}{!}{
  \begin{tabular}{l|c|c|c|l}
    \toprule
    & \multicolumn{2}{c}{\textbf{Policy Representation}} & \multicolumn{2}{c}{\textbf{Request Evaluation}} 
    \\\hline
    \textbf{Domain Policy Construct} & \textbf{Relevant} & \textbf{Implemented} & \textbf{Relevant} & \textbf{Implemented} 
    \\\hline
    \ \ Provision           & yes  & yes     & yes & yes        \\\hline
    \ \ Parsed logical rule & yes  & yes     & no  & no         \\\hline
    \ \ Obligation          & yes  & partial & yes & partial    \\\hline
    \textbf{Requesters} \\ \hline
    \ \ Device              & yes  & yes     & yes & yes        \\\hline
    \ \ Organization        & yes  & yes     & yes & yes        \\\hline
    \ \ Dependency          & yes  & uc      & uc  & uc         \\\hline
    \ \ Licensee            & yes  & partial & yes & partial    \\\hline
    \textbf{Affiliations} \\ \hline
    \ \ Federal/Non-Federal & yes  & partial & yes & partial    \\\hline
    \ \ Named requester     & yes  & yes     & yes & yes        \\\hline
    \textbf{Frequencies} \\ \hline
    \ \ Frequency range     & yes  & yes     & yes & yes        \\\hline
    \ \ Single frequency    & yes  & yes     & yes & yes        \\\hline
    \ \ Named bands         & yes  & yes     & uc  & uc         \\\hline
    \ \ Units               & yes  & yes     & yes & yes        \\\hline
    \textbf{Time} \\ \hline
    \ \ Timezones           & yes  & yes     & yes & yes        \\\hline
    \ \ Policy validity     & yes  & yes     & yes & yes        \\\hline
    \textbf{Locations} \\ \hline
    \ \ Named locations     & yes  & yes     & yes & yes        \\\hline
    \ \ Relative locations  & yes  & uc      & yes & uc         \\\hline
    \ \ Polygons/Circles    & yes  & yes     & yes & yes        \\\hline
    \textbf{Geographical rules} \\ \hline
    \ \ Specific location   & yes  & yes     & yes & yes        \\\hline
    \ \ List of locations   & yes  & yes     & yes & yes        \\\hline
    \textbf{Precedence} \\ \hline
    \ \ Levels              & yes  & yes     & yes & yes        \\\hline
    \textbf{Explanations} \\ \hline
    \ \ Policy triggered    & yes  & yes     & yes & yes        \\\hline
    \ \ Rules not satisfied & yes  & yes     & yes & yes        \\
    \bottomrule
\end{tabular}}
\caption{DSA Policy Framework policy semantics coverage}
\label{tab:evaluation}
\end{table}

The basic structure of a policy varies among source documents. The NTIA Redbook uses provisions to distinguish policy behavior with regards to attributes (specific device or location, for instance). Provisions are described as sub-policies in the policy capture spreadsheet and in the Framework, and they are leveraged during request evaluation. Policies can be specified in the \emph{Parser logical rule} format, which is enabled in the Policy Builder. The framework does not currently represent the individual elements of an obligation. Instead, it is represented as text or as a canned identifier to support request evaluation. 

The text in the source policies specifies regulations for a variety of requester types. While all requester types are relevant for policy representation, requests only express the device. However, the DSA KG does represent several types including Organization and Licensee, so if a request was received for one of these types, theoretically, it could be evaluated. If a policy specifies a dependency between requesters, this dependency is only currently treated as text. Policies can also regulate the spectrum usage by affiliation (Federal systems are permitted to use some frequency, for instance). The Framework allows affiliation rules to be specified in policies, however, affiliation reasoning is currently limited by the expressiveness of affiliations in the DSA Ontology, where only requesters with an explicit affiliation (named requester) are effectively reasoned.

Frequency rules are specified in terms of a range or single frequency in different units (MHz, GHz). Sometimes, a range is described as a band, which is a named frequency range. Requests do not express named bands. The framework supports all of these constructs, using the \mytexttt{Frequency} and \mytexttt{FrequencyRange} attributes, and the \mytexttt{sio:hasUnit} property.  The time attribute exists in local policies only, and they specify the validity of a policy. Time and timezones are supported using \mytexttt{xsd:dateTime} literals and PROV-O time predicates.

Most policies refer to locations by names or by coordinates (points, polygons, and circles), but sometimes a location is expressed in relation to another location. The framework uses Census.gov shapes to refine named locations and WKT literals to represent polygons and circles. Relative locations are still under development. Geographical rules are defined in terms of the requester being in a location or in a list of locations. The framework implements all of these constructs using \mytexttt{geo:sfWithin} predicate and OWL unions.

The framework implements all of the identified precedence needs. It can define and evaluate precedence levels. For explaining evaluation results, the framework implements all current explanation requirements. It outputs which policy was triggered by a request and presents reasons based on the presented heuristics. Finally, the constructs identified as ``uc'' are areas for future work.

\section{System Adoption and Deployment}

The DSA Policy Framework is being used in simulated scenarios, where it supports the research \& development of other components of a dynamic spectrum management system. It currently contains approximately 165 high-level policies from the NTIA Redbook (including their sub-policies). The DSA Ontology contains 695 classes and is constantly evolving to address new domain constructs and support more precise request evaluation.

The framework is transitioning to support live, over-the-air field exercises involving a diverse set of federal and commercial radios. During these exercises, the Framework supports (1) the creation, deletion, and revision of local policies, (2) the real-time processing of numerous spectrum requests, and (3) the generation of explanations that describe how the spectrum requests were processed. The public released assets developed during the course of the project can be accessed at \url{https://github.com/tetherless-world/dsa-open/}.

\section{Related Work}

Kirrane~\cite{kirrane_access_2017} offers a comprehensive survey of access control models, well-known policy languages, proposed frameworks that utilize ontologies and/or rules to express policies, and a categorization of policy languages and frameworks against access control requirements. XACML 3.0, the eXtensible Access Control Markup Language~\cite{noauthor_extensible_nodate}, is a well-known policy language and \textit{de facto} standard for representing attribute-based access control (ABAC)~\cite{hu_abac_2015} policies and requests. Importantly, XACML provides a reference architecture for centralizing access control and a process model for evaluating requests against existing policies that inform the design of access control systems across domains and technologies.

Thi~\cite{tran_thi_x-strowl_2012} proposes an OWL-based extension to XACML to support a generalized context-aware role-based access control (RBAC) model providing spatio-temporal restrictions and conforming with the NIST RBAC standard~\cite{ferraiolo_rbac_2009}. Their work augments the XACML architecture with new functions and data types.

Muppavarapu~\cite{muppavarapu_semantic-based_2008} identifies the limitations of identity-based access control schemes used in the Open Grid Services Architecture (OGSA) and proposes the use of OWL to represent the ontology of an organization’s resources and users. They further propose the use of semantics in conjunction with the XACML standard for better interoperability and reduced administration overhead.

Our approach combines OWL, PROV-O, and the HermiT OWL reasoner with an ontology, represented as a knowledge graph, to support the representation of policies governing access to available spectrum.  Relevant related research is described in Dundua~\cite{dundua_towards_2019}, where previous work is described that uses OWL for modeling and analyzing access control policies, especially ABAC, and considers how the ABAC model can be integrated into ontology languages. In addition, Sharma~\cite{sharma_representing_2016} describes how OWL can be used to formally define and process security policies that can be captured using ABAC models. This work demonstrates how models, domains, data and security policies can be expressed in OWL and how a reasoner can be used to decide what is permitted.

Kolovski~\cite{kolovski_representing_2005} maps the web service policy language, WS-Policy~\cite{ws-policy}, to the description logic fragment species of OWL and demonstrates how standard OWL reasoners can check policy conformity and perform policy analysis tasks.

Garcìa~\cite{garcia_web_2007,garcia_semantic_2013} and Finin~\cite{finin_rowlbac_2008} offer important contributions on how end-to-end usage rights and access control systems may be implemented in OWL and RDF. Garcìa proposes a ``Copyright Ontology'' based on OWL and RDF for expressing rights, representations that can be associated with media fragments in a web-scale ``rights value change.'' Finin describes two ways to support standard RBAC models in OWL and discusses how their OWL implementations can be extended to model attribute-based RBAC or, more generally, ABAC. 

Our policy representation approach builds on previous work by matching the cross-domain policy expression semantics of XACML.  It extends these semantics with the capacity to express rich spatio-temporal restrictions, enabling the implementation of a wide variety of attribute-based policies across domains.  It leverages background knowledge from domain-specific knowledge graphs that are structured with a domain-derived ontology, enabling the inference of policy applicability based on attributes and constraints. Our approach uniquely conceptualizes policy requests as PROV activities and request evaluations as realizations. Finally, our approach provides a novel reasoner-based explanation in request evaluation results, enabling domain policy developers to understand the precise reasons for policy decisions. 

\section{Conclusion}

We described a policy Framework that leverages a machine-readable, radio spectrum policy representation to support policy management and enable a domain-specific reasoner to efficiently process spectrum requests. The DSA policy model uses OWL restrictions on PROV-O properties to represent policy rules in a hierarchical approach that maximizes the reuse of rules when local policies are created, therefore facilitating the creation of local policies. The hierarchical nature of this policy representation also supports the explanation of evaluation results, by traversing the graph to find rules that were not satisfied. Because it leverages a domain Knowledge Graph, built from multiple knowledge sources, the domain-specific reasoner allows rich semantics which otherwise would be difficult to achieve with approaches that rely on a ``flat'' representation of attributes.

During the course of the project, we encountered OWL reasoning performance issues when multiple requests are simultaneously received. To decrease reasoning time, we partitioned the request graph into smaller graphs and evaluated each in parallel, using multiple processor cores. This allowed us to match the required response time of under 10 seconds.

Future work includes additional support for enforcement of the DSA policy model. Although the DSA policy model's OWL hierarchy maximizes the reuse of rules, there is currently no enforcement. Overlapping rules can be created, which can lead to multiple policies with the same precedence level being triggered during request evaluation. The DSA Ontology is constantly changing as additional policies are added to the framework with terms that have yet to be defined. Existing policies can be affected by changes in the DSA Ontology as their rules reference entities in it and they may need to be reviewed with regards to the updated representation of the domain. We plan to provide spectrum managers a way of tracking policies that are subject to review due to ontology changes.

\bigskip\noindent\textbf{Acknowledgement of Support and Disclaimer.}
This work is funded in support of National Spectrum Consortium (NSC) project number NSC-17-7030. Any opinions, findings and conclusions or recommendations expressed in this material are those the authors and do not necessarily reflect the views of AFRL.

\bibliographystyle{splncs04}
\bibliography{bib}

\end{document}